\documentclass{article}

\usepackage[preprint,nonatbib]{neurips_2024}

\usepackage[numbers]{natbib}
\usepackage[utf8]{inputenc} % allow utf-8 input
\usepackage[T1]{fontenc}    % use 8-bit T1 fonts
\usepackage{hyperref}       % hyperlinks
\usepackage{url}            % simple URL typesetting
\usepackage{booktabs}       % professional-quality tables
\usepackage{amsfonts}       % blackboard math symbols
\usepackage{nicefrac}       % compact symbols for 1/2, etc.
\usepackage{microtype}      % microtypography
\usepackage{xcolor}         % colors
\usepackage{inconsolata}
\usepackage{algorithm}
\usepackage{algorithmic}
\usepackage{booktabs}
\usepackage{tabularx}
\usepackage{graphicx}
\usepackage{amsmath}
\usepackage{amssymb}
\usepackage{bbding}
\usepackage{makecell}
\usepackage{wrapfig}
\usepackage{booktabs}
\usepackage{mathrsfs}
\usepackage{multirow}
\usepackage{xcolor}
\usepackage{colortbl}
\usepackage{pifont}
\usepackage{paralist}
\newcommand{\grayrule}{\arrayrulecolor{gray}\midrule\arrayrulecolor{black}}

\newcommand\modelname{\textit{AlchemistCoder}}
\newcommand\promptname{\textit{AlchemistPrompt}}

\makeatletter
\newcommand\figcaption{\def\@captype{figure}\caption}
\newcommand\tabcaption{\def\@captype{table}\caption}
\makeatother

\title{AlchemistCoder: Harmonizing and Eliciting Code Capability by Hindsight Tuning on Multi-source Data}

\author{Zifan Song$^{1,2*}$ \quad Yudong Wang$^{2}$\thanks{\quad Equal contributions} \quad Wenwei Zhang$^{2*}$ \quad Kuikun Liu$^{2}$ \\
\textbf{Chengqi Lyu$^{2}$ \quad Demin Song$^{2}$ \quad Qipeng Guo$^{2}$ \quad Hang Yan$^{2}$} \\
\textbf{Dahua Lin$^{2,3}$ \quad Kai Chen$^{2 \dag}$ \quad Cairong Zhao$^{1}$\thanks{\quad Corresponding author}}\\
$^{1}$Tongji University
$^{2}$Shanghai AI Laboratory\\
$^{3}$Chinese University of Hong Kong\\
}

\begin{document}

\maketitle

\begin{abstract}
Open-source Large Language Models (LLMs) and their specialized variants, particularly Code LLMs, have recently delivered impressive performance. However, previous Code LLMs are typically fine-tuned on single-source data with limited quality and diversity, which may insufficiently elicit the potential of pre-trained LLMs. In this paper, we present \textbf{\textit{AlchemistCoder}}, a series of Code LLMs with enhanced code generation and generalization capabilities fine-tuned on multi-source data. To achieve this, we pioneer to unveil inherent conflicts among the various styles and qualities in multi-source code corpora and introduce data-specific prompts with hindsight relabeling, termed \textbf{\textit{AlchemistPrompts}}, to harmonize different data sources and instruction-response pairs. Additionally, we propose incorporating the data construction process into the fine-tuning data as code comprehension tasks, including instruction evolution, data filtering, and code review. Extensive experiments demonstrate that \textit{\textbf{AlchemistCoder}} holds a clear lead among all models of the same size (6.7B/7B) and rivals or even surpasses larger models (15B/33B/70B), showcasing the efficacy of our method in refining instruction-following capabilities and advancing the boundaries of code intelligence. Source code and models are available at \url{https://github.com/InternLM/AlchemistCoder}.
\end{abstract}

\section{Introduction}
% \begin{figure}[h]
%     \begin{minipage}[]{0.47\textwidth}
%     \end{minipage}
%     \hfill%
%     \begin{minipage}[]{0.5\textwidth}
%             \includegraphics[width=1.00\columnwidth]{figure/figure_comparison_on_humaneval_and_mbpp.pdf}
%             \vspace{-10pt}
%             \caption{Performance scatter plot (\textit{top right} is better) of open-source models on mainstream code benchmarks, HumanEval and MBPP. Our \modelname{} series demonstrates astonishing performance across all open-source Code LLMs.}
%             \label{figure: Comparison on HumanEval and MBPP}
%             \vspace{-12pt}
%     \end{minipage}
% \end{figure}

\begin{wrapfigure}{r}{0.5\textwidth}
\vspace{-50pt}
\centering
\includegraphics[width=0.5\textwidth]{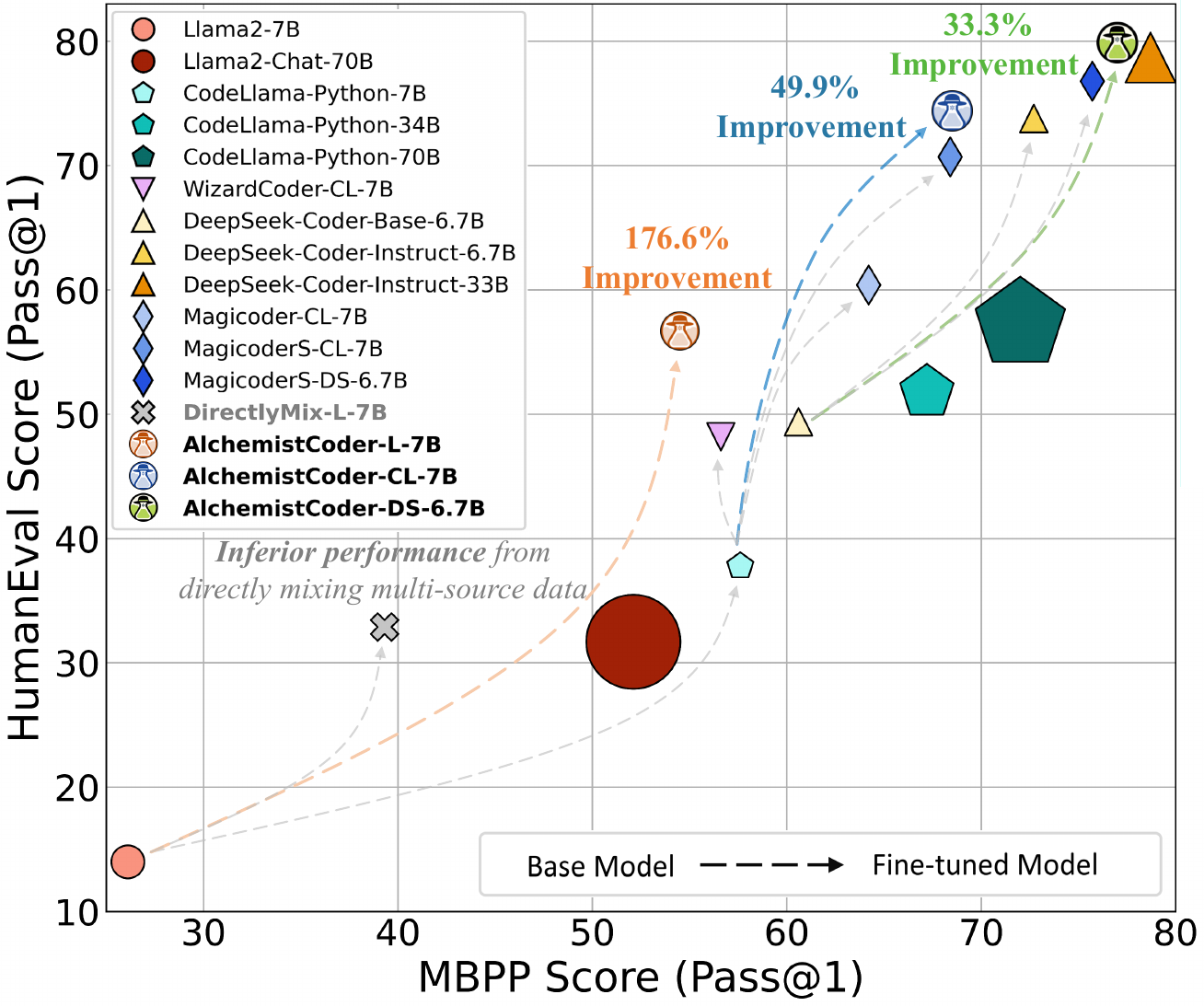}
\vspace{-17pt}
% \caption{{Performance scatter plot (\textit{top right} is better) of open-source models on mainstream code benchmarks, HumanEval and MBPP.} We use different shapes/sizes to represent different series/parameters of models, and draw dashed lines to indicate the improvements of the fine-tuned Code LLMs compared to the base models. -L/-CL/-DS respectively indicate the models fine-tuned on Llama 2/CodeLlama-Python/DeepSeekCoder-Base.}
\caption{Performance scatter plot (\textit{top right} is better) of open-source models on mainstream code benchmarks, HumanEval and MBPP. Our \modelname{} series demonstrates astonishing performance across all open-source Code LLMs.}
\label{figure: Comparison on HumanEval and MBPP}
\vspace{-50pt}
\end{wrapfigure}

Closed-source Large Language Models (LLMs) like ChatGPT and GPT-4 \cite{chatgpt, GPT-4} exhibit impressive code intelligence by learning on large-scale and diverse code corpus, which also benefits many other applications, such as math reasoning~\cite{pot}, embodied control~\cite{code_as_policy}, and agent~\cite{agent_code_survey}.
% excellent performance on coding tasks, but also
Since open-source LLMs~\cite{llama2} still lag behind closed-source LLMs~\cite{GPT-4} in this field, there has been growing interest in investigating the acquisition of code capabilities by developing specialized Code LLMs~\cite{codellama, deepseek-coder}.

The training of Code LLMs mainly goes through pre-training and fine-tuning stages \cite{codellama}.
Pioneer works \cite{humaneval, nijkamp2022codegen, starcoder} have amassed extensive code data for pre-training, while recent open-source models \cite{wizardcoder, wei2023magicoder} highight the effectiveness of high-quality or targeted code fine-tuning datasets.
Despite these advancements, current fine-tuning methods mainly rely on a specific type of code-related question-answering dataset, unlike the pre-training stage that integrates code-related corpus from various sources \cite{codellama}. Such a discrepancy indicates that the fine-tuning data may lack the necessary diversity to fully stimulate the capabilities of base models, resulting in limited performance, generalization, and robustness.

% \zifan{Despite these advancements, existing instruction fine-tuned Code LLMs are typically trained on specific datasets with a single data style, leading to the potential for models to fall into limited domains and even requiring specialized language guidance to perform in a constrained manner. Additionally, this results in existing open-source datasets for instruction tuning being like scattered grains of sand, not effectively accumulated into a pagoda. These highlight the ongoing need for innovative strategies of integrating multi-source data to advance the boundaries of code intelligence.}
% Despite these advancements, pre-trained Code LLMs have mixed multi-source data and performed macro-level data processing (\textit{e.g.}, data cleaning and data scaling) but rarely pay attention to the optimization of the data itself. Instruction-tuned models that integrate multi-source data for training are few and far between, typically training on a single specific dataset with limited data style. This may result in the potential for fine-tuned Code LLMs to fall into limited domains and perform in a constrained manner, highlighting the ongoing need for innovative strategies of integrating multi-source data that vary in style and quality for instruction tuning. 
% \yudong{Do we need to highlight the difference between open-source and closed-source models? How about highlighting a single data style will cause the model to fall into a certain domain, causing instructions to be unable to follow and performance to be reduced, balaba... }

\begin{figure*}[t]
\centering
\includegraphics[width=0.98\textwidth]{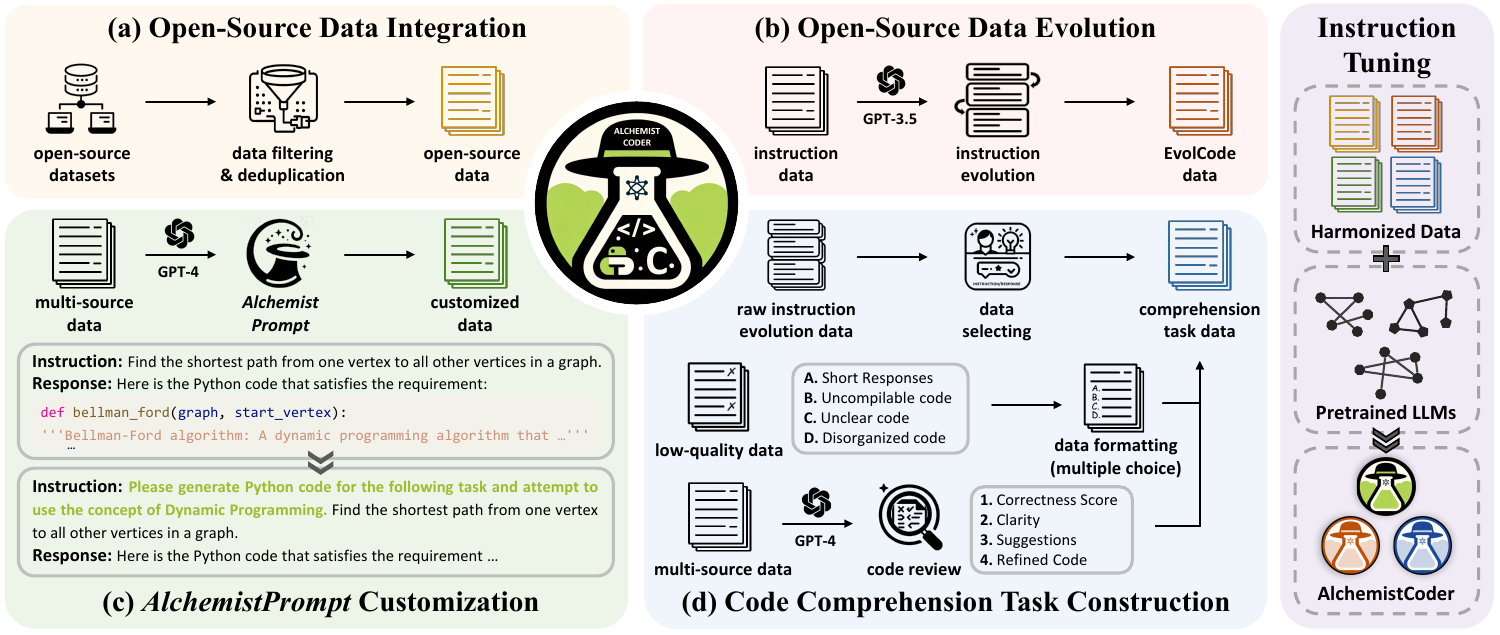}
\vspace{-3pt}
\caption{Overview for developing \textit{AlchemistCoder} series. We first integrate high-quality open-source data (a) and conduct data evolution based on them (b). Then, we adopt \promptname{} to harmonize their inherent conflicts (c) and construct code comprehension data (d). We use a mix of these data to fine-tune various pre-trained LLMs to obtain our \modelname{} models.}
\label{figure: Overview}
\vspace{-7pt}
\end{figure*}

To overcome the limitations in quality and diversity within single-source data, we pioneer to explore integrating multi-source data for Code LLM fine-tuning. However, this is a non-trivial paradigm and blindly integrating multi-source data can potentially lead to inferior performance (\textit{e.g.}, the DirectlyMix-L-7B model in Fig. \ref{figure: Comparison on HumanEval and MBPP}). To track this, we unveil inherent conflicts in multi-source code corpora, including conflicting code language requirements and response styles. 
% \yudong{seems the figure cannot express $\mathrm{1+1>2}$}
Inspired by hindsight relabeling \cite{hindsight_openai_2017, hindsight_wisdom_2023}, we propose to design data-specific prompts to harmonize the inherent conflicts for multi-source data integration, better eliciting the performance of base models.
We term this form of prompts as \promptname{}s, inspired by the power and definition of \textit{Alchemists}:
% \vspace{-2pt}
\begin{quote}
\small
\textit{``Alchemist: Someone Who Transforms Things for the Better.''}    —— 
 Merriam Webster 
\end{quote}
% \vspace{-2pt}
Specifically, we first integrate several open-source code datasets and conduct instruction evolution~\cite{wizardcoder} based on some of them (Fig.~\ref{figure: Overview}(a, b)).
% Due to the diverse styles, coding languages, and varying data quality of different sources, directly mixing multi-source data will cause the trained model to be unaligned to specific coding languages and styles, resulting in inferior performance.
As shown in Fig.~\ref{figure: Overview}(c), for instruction-response pairs of different sources, we adopt one LLM to generate \promptname{}s that accurately and explicitly describe the characteristics as requirements of the response to enrich the instructions.
In-depth, the efficacy of \promptname{}s is twofold: 1) Harmonization between different data sources: \promptname{}s generated from the same LLM have similar styles and can bridge the style differences between sources, while the introduction of \promptname{}-customized data, accounting for only 5\%, achieves a balance between data diversity and domain gaps;
2) Harmonization within instruction-response pairs: As fine-grained and data-specific prompts, \promptname{}s are designed to augment instructions with specific programming languages, algorithm concepts, and other code-related information involved in responses, which can refine the alignment within instruction-response pairs and enhance the instruction-following abilities of fine-tuned models.

% This makes the response more consistent with and goal-conditioned on the new instructions in each data source and eventually transforms the fine-tuning process on multi-source data from \textit{cloning different responses to similar questions} to \textit{learning to follow diverse instructions}.
% Consequently, \promptname{}s not only enhance the instruction-following capability of LLMs but also allow the integrated multi-source data to effectively boost different aspects of the base models.
% To track these challenges, we propose data-specific \promptname{} to polish multi-source data at a fine-grained level, enhancing the condition/goal of data from a hindsight perspective \cite{hindsight_openai_2017, chain_of_hindsight} to refine the instruction-following capabilities of models. Simultaneously, \promptname{}s generated from the same model can bridge the domain-gap between various types of data, effectively boosting the benefit from the integration of multi-source data. Specifically, we adopt a LLM (GPT-4 in our experiments) as the ALCHEMIST to examine instruction/response paired samples and generate appropriate preceding descriptions to optimize instructions, thereby achieving a better alignment between instructions and responses. 

Apart from the conventional problem-solution data, we argue that the progression of code data (\textit{e.g.}, data evolution, cleaning, and quality evaluation) reflects higher-level capabilities and offers valuable insights for the enhancement of Code LLMs. Consequently, we delineate the construction of data into three integral tasks for training: instruction evolution, data filtering, and code review (see Fig. \ref{figure: Overview} (d)), facilitating enhanced code comprehension capabilities.

We conduct extensive experiments with various base models \cite{llama2, codellama, deepseek-coder} and develop the instruction-tuned \modelname{} series. As shown in Fig. \ref{figure: Comparison on HumanEval and MBPP}, on two mainstream code benchmarks, HumanEval and MBPP, \modelname{} holds a clear lead among all models of equivalent size (6.7B/7B), and rivals or even surpasses larger models (15B/33B/70B), demonstrating harmonized and formidable code capabilities.
Furthermore, we delve into the effectiveness of \promptname{}s and discern that they alleviate the misalignment between instructions and responses within the data.
Remarkably, \promptname{}s allow the code corpus to also significantly improve the general capability of Code LLMs, as demonstrated by the improvements on MMLU, BBH, and GSM8K.
% \zifan{Simultaneously, our models attain significant improvement on non-code tasks (\textit{e.g.}, MMLU and BBH), bridging the gap between code capabilities and other abilities seen in previous perspectives \cite{xu2023lemur}.}
Our main contributions are summarized as follows:
\begin{itemize}
\item Our work pioneers to integrate multi-source data for Code LLM fine-tuning to overcome the limitations of quality and diversity inherent in single-source data. 
\item We unveil inherent conflicts within multi-source code corpora and introduce \promptname{}s, revealing the power of hindsight tuning for code generation, aiming to harmonize the conflicts among sources and bridge the alignment within instruction-response pairs. 
% \item \promptname{}s are introduced to reveal the power of hindsight tuning in the context of code generation, aiming to harmonize multi-source data and bridge the alignment between instructions/responses of data. 
\item We propose to incorporate data construction process into the fine-tuning data and design code comprehension tasks, including instruction evolution, data filtering, and code review, endowing comprehensive code capabilities.
\item Extensive ablation and analytical studies confirm the efficacy of our key concepts for enhancing the diversity, quality, and cost-effectiveness of Code LLM fine-tuning data. Through instruction tuning on various base models, we develop the \modelname{} series, surpassing all Code LLMs of the same size on a wide spectrum of code benchmarks.

% verifying the universality and efficacy of our method to further the performance of the base models.
\end{itemize}

\section{Method}
To more comprehensively elicit the capability of the base LLMs, we first construct multi-source data for fine-tuning (\S~\ref{section:multi_source_data}), which is harmonized by \promptname{}s to take effect(\S~\ref{section: AlchemistPrompt}).
Code comprehension tasks are also constructed to further improve the performance(\S~\ref{section:code_comprehension}). We also discuss the details and statistics of the filtered and harmonized multi-source data in \S~\ref{section:data_clean}.

\begin{figure}[t]
\centering
\includegraphics[width=0.98\textwidth]{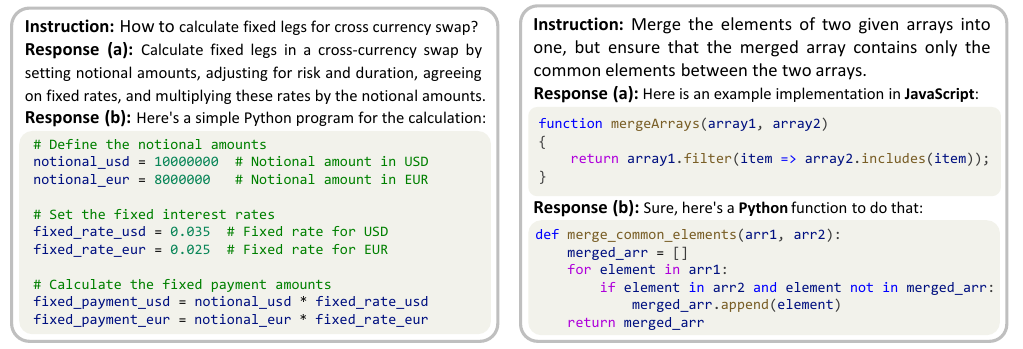}
\vspace{-5pt}
\caption{Examples of conflicts (\textit{e.g.}, various styles and quality) within multi-source code corpora. }
\label{figure: Inherent Conflict}
\vspace{-7pt}
\end{figure}

\subsection{Multi-source data construction}\label{section:multi_source_data}
To fully elicit the capability of code LLMs, we first collect the fine-tuning data from multiple sources (Fig.~\ref{figure: Overview}(a)) and adopt the instruction evolution~\cite{wizardcoder} to improve the complexity of the instructions (Fig.~\ref{figure: Overview}(b)).
However, integrating multi-source data for instruction tuning poses challenges.
% and previous pre-trained open-source models, such as CodeLlama \cite{codellama} and LEMUR \cite{xu2023lemur}, have explored merging various types of data for training. Despite conducting macro-level data processing (\textit{e.g.}, data cleaning, filtering, and scaling in data proportions), these works do not meticulously attend to the refinement of the data content itself. In the instruction tuning phase, Code LLMs that integrate multi-source data for training are few and far between. 
Naturally, one code-related question can be solved by different coding languages with various algorithms or response styles (\textit{e.g.}, with or without reasoning). When naively combing data curated by different developers with different LLMs, the model tends to learn to answer similar questions with different coding languages and response styles, as depicted in Fig. \ref{figure: Inherent Conflict}.
On the one hand, learning diverse responses may elicit different capability aspects of the base models.
On the other hand, since the learned responses to similar instructions often diverge due to implicit human intentions, the LLMs tend to be unaligned (to our expectation) after the fine-tuning on the directly mixed data (\textit{e.g.}, we cannot expect which coding language the LLMs will use in real-world applications), resulting in inferior performance.
Therefore, directly mixing multi-source data is not a promising solution and can be detrimental. 

\begin{wrapfigure}{r}{0.5\textwidth}
\vspace{-10pt}
\centering
\includegraphics[width=0.5\textwidth]{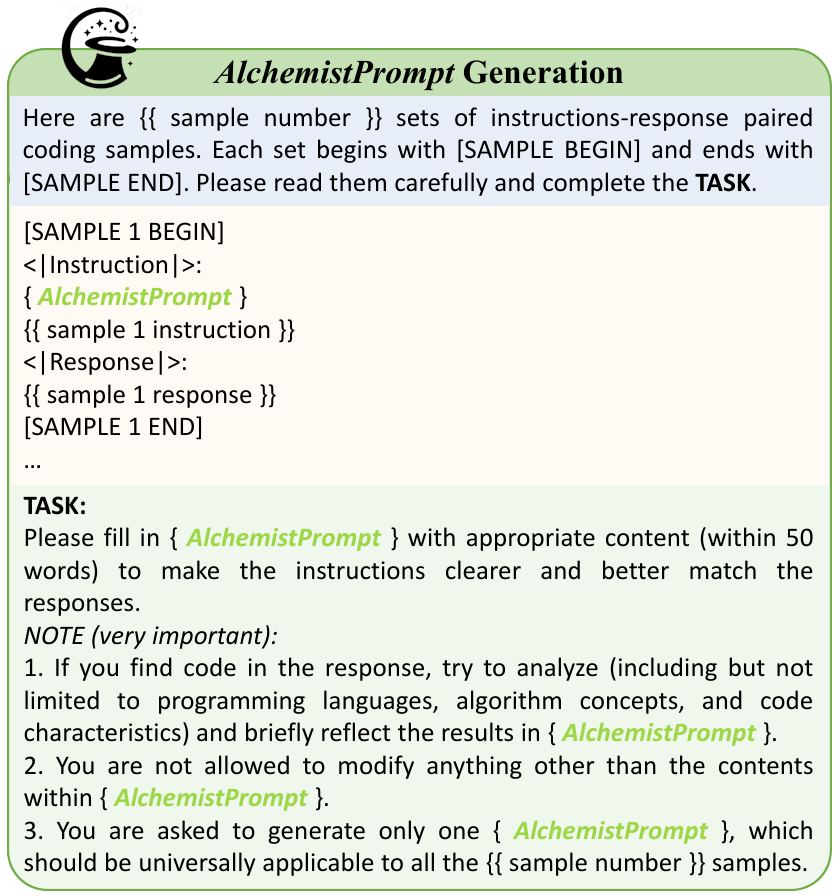} % Reduce the figure size so that it is slightly narrower than the column.
\vspace{-12pt}
\caption{Detailed prompt designed for generating data-specific \promptname{}s. }
\label{figure: AlchemistPrompt Generation}
\vspace{-7pt}
\end{wrapfigure}

\subsection{AlchemistPrompt}
\label{section: AlchemistPrompt}
% The currently present training data is sourced from diverse origins, exhibiting variations in data types and quality. Simultaneously, a considerable disparity exists in the style of responses and evaluation methods. 

% If we liken learning models from data to humans absorbing nutrients from food, then pre-trained models are akin to stewing all ingredients in a big pot, whereas fine-tuning models is like tasting only one dish, the former leading to nutrient loss and the latter to nutrient imbalance. Moreover, just as every diner has their own preferences for dishes, a personalized restaurant would inquire about each customer's favored cooking methods for each dish to cater to their tastes. Similarly, different models might show ``understanding'' and ``preferences'' for various data, demonstrating a ``preference'' for data that matches their ``taste''.
To harmonize the inherent conflicts within multi-source data, we propose to customize data-specific prompts called \promptname{}s,  (Fig.~\ref{figure: Overview}(c)), inspired by the concept of hindsight~\cite{hindsight_openai_2017, hindsight_wisdom_2023}.
Specifically, we employ GPT-4~\cite{GPT-4} to play the role of an \textit{Alchemist} and design the prompt as illustrated in Fig. \ref{figure: AlchemistPrompt Generation} to obtain \promptname{}s. 
For instance, for an instruction of `Write code to find the shortest path from one vertex to all other vertices in a graph', if the response involves Python code of a Bellman-Ford algorithm with dynamic programming, we would expect to customize the instruction with an \promptname{} of `Please generate Python code for the following task and attempt to use the concept of Dynamic Programming'.

The adjustments to data made by \promptname{}s are relatively minor and well-calibrated. Our ablation study indicates that the optimal performance can be achieved by incorporating \promptname{}s into only 5\% of all the samples, striking a balance between the diversity and domain gap resulting from the fusion of multi-source data. Crucially, by retrospectively analyzing previous responses and reinterpreting them as alternate goals, the \promptname{}s serve to elevate the condition/goal of the data. This hindsight integration \cite{hindsight_openai_2017, hindsight_wisdom_2023, chain_of_hindsight} allows for a more nuanced and adaptive learning process, enhancing not only the models' comprehension of data but also refining their instruction-following capabilities.

\subsection{Code comprehension task}\label{section:code_comprehension}
The existing training datasets for Code LLMs \cite{alphacode, codealpaca, evolcodealpaca, wizardcoder, wei2023magicoder} primarily focus on the code generation task consisting of programming problems and their corresponding code solutions. However, we contend that beyond this, the process of constructing code data demonstrates higher-level abilities. Consequently, we devise three code comprehension tasks relevant to data construction, including instruction evolution, data filtering, and code review (Fig.~\ref{figure: Overview}(d)). 

\noindent\textbf{{Instruction evolution.}}
Inspired by the concept of instruction evolution \cite{xu2023wizardlm, wizardcoder}, we employ GPT-3.5 \cite{chatgpt} to construct instruction evolution task data, which entails augmenting the requirements for instructions and providing detailed explanations for programming tasks. Integrating the instruction evolution task aids the model in discerning the disparities before and after evolutions, thereby deepening the comprehension of programming requirements, code complexity, task decomposition, and other code-related concepts.

\noindent\textbf{{Data Filtering.}}
We identify four categories of low-quality data from multiple sources: (a) responses that are excessively short and lack code, (b) code that fails to compile, (c) code with poor clarity, and (d) code that does not adhere to the requirement in the instruction regarding its organization in function form. Each instruction in the data filtering task presents the model with a low-quality sample and prompts the model to classify it into one of the four categories. The data filtering task entails recycling the filtered-out data by offering counterexamples, thereby assisting the model in generating fewer low-quality responses.

\noindent\textbf{{Code review.}}
In this task, we require the model to review a piece of code and assign scores between 0 and 10 for correctness and clarity separately. Additionally, the model is expected to provide suggestions for code improvement and present the refined code. To obtain higher-quality data, we utilize GPT-4 \cite{GPT-4} to generate code reviews and select cases that are more representative, particularly those with average correctness and clarity scores exceeding 8 or falling below 6. Simultaneously, we focus on instances where one aspect exhibits severe deficiencies, \textit{i.e.}, the score of correctness or clarity is equal to or below 4.

\subsection{Data cleaning and decontamination}\label{section:data_clean}
In practice, we have established a set of filtering rules to enhance our data cleaning and purification procedures. These rules involve excluding samples based on various criteria, such as response length (either too short or too long), absence of code or insufficient code content, non-compilable code, code failing test cases (pertinent to certain samples), responses structured in notebook form, and instances with excessive textual descriptions preceding the code. After conducting an extensive series of validation experiments, we conclusively decide to eliminate low-quality data meeting either of the following conditions: (a) responses that are excessively brief and lack code. Such responses typically offer direct answers to the instructions, neglecting both the code solution and explanatory annotations. Additionally, these samples frequently present overly simplistic questions in the instructions; (b) code solutions that are non-compilable or fail test cases (relevant to specific samples).

Concurrently, following \cite{gunasekar2023textbooks}, we employ N-gram similarity, cosine distance of code embeddings, and edit distance of code syntax trees to calculate the similarity between training data and samples in HumanEval and MBPP. We subsequently discard samples through this process of data filtering and deduplication, resulting in the removal of approximately 6\% of the dataset.

\subsection{Harmonized AlchemistCoder dataset}\label{section:alchemistcoder_dataset}
Our \textit{AlchemistCoder} dataset ($\sim$200M tokens) comprises four types of multi-source data, encompassing open-source datasets and three types of data constructed by us. Specifically, (a) open-source datasets including Evol-Instruct-Code-80k-v1 \cite{evolinstructcode}, CodeExercise-Python-27k \cite{codeexercise}, and evol-codealpaca-v1 \cite{evolcodealpaca}, (b) EvolCode data generated from gpt-3.5-turbo following \cite{wizardcoder}, (c) data customized by \promptname{}s, and (d) data of the code comprehension tasks (\textit{i.e.}, instruction evolution, data filtering, and code review).

\begin{figure}[t]
    \centering
    \begin{minipage}{0.48\textwidth}
        \centering
        \includegraphics[width=0.93\textwidth]{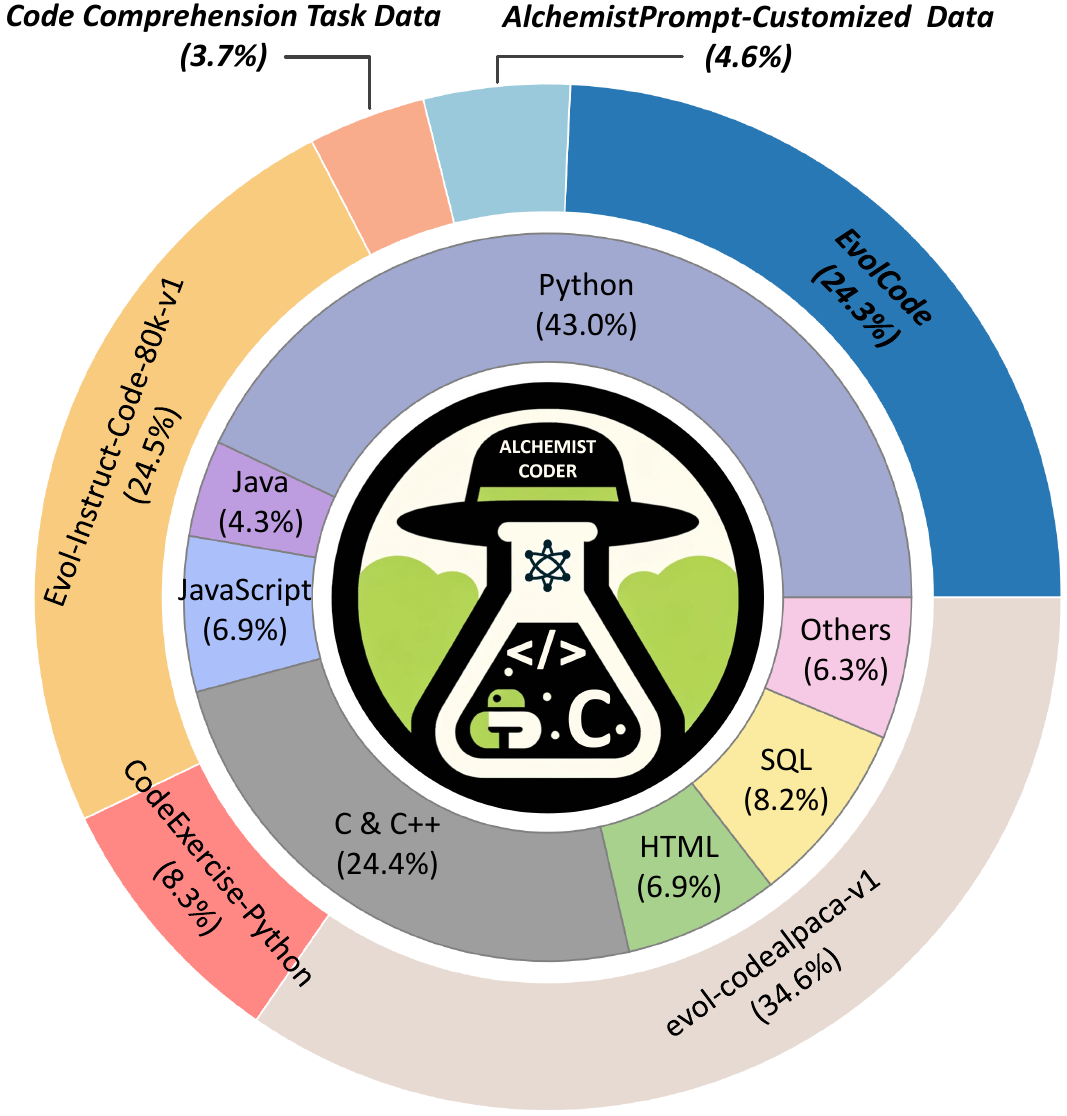}
        \vspace{-5pt}
        \caption{Data distribution analysis of our \textit{AlchemistCoder} dataset. The outer and inner circular diagrams respectively display the distributions of data composition and programming languages. Data from \promptname{}s and code comprehension tasks, constituting only 8\% of the total data, plays a crucial role in harmonizing and polishing the fine-tuning data.
        }
        \label{figure: Data Distribution}
    \end{minipage}
    \hfill
    \begin{minipage}{0.48\textwidth}
        \centering
        \includegraphics[width=0.93\textwidth]{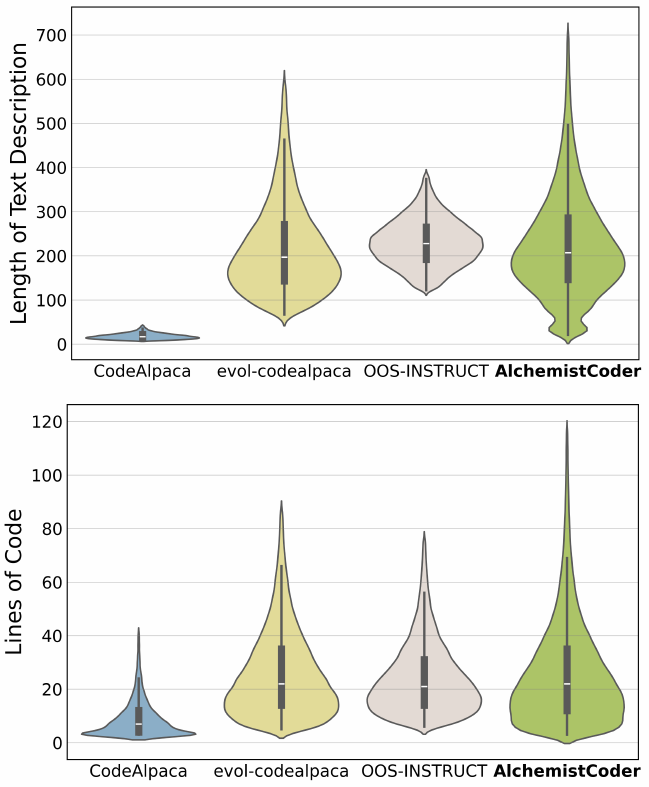}
        \vspace{-5pt}
        \caption{Comparative distribution of text description lengths (top) and code lines (bottom). Our dataset contains high-quality samples with more diverse distributions.}
        \label{figure: Code and Description Length}
    \end{minipage}
\vspace{-12pt}
\end{figure}

We visualize the distributions of data sources and programming languages  using two circular graphs in Fig. \ref{figure: Data Distribution}. Concurrently,  Fig. \ref{figure: Code and Description Length} reports a distribution of text description lengths and code lines. Compared to CodeAlpaca \cite{codealpaca} and OOS-INSTRUCT \cite{wei2023magicoder}, our \textit{AlchemistCoder} dataset presents a notably diverse distribution and maintains moderate overall text description and code lengths, benefiting significantly from the integration of multi-source data along with \promptname{}s and code comprehension tasks. This is instrumental in contributing to a comprehensive evolution of code capability.

\section{Experiments}
In this section, we report results on various benchmarks of code generation and conduct ablation experiments. Furthermore, we present analytical studies to provide a more in-depth demonstration of the efficacy of our \textit{AlchemistCoder}.

\subsection{Benchmarks and implementation details}\label{Benchmarks and implementation details}
\noindent\textbf{{Benchmarks.}}
We adopt six code benchmarks: HumanEval \cite{humaneval}, HumanEval+ \cite{humaneval+}, HumanEval-X \cite{humanevalx}, MBPP \cite{mbpp}, MBPP+ \cite{humaneval+}, and DS-1000 \cite{ds-1000}. 
% Following prior works \cite{humanevalx, chen2023teaching}, we use the greedy decoding strategy and focus on comparing the pass@1 metric. 
In addition, we access three mainstream benchmarks (MMLU \cite{mmlu}, BBH \cite{bbh}, and GSM8K \cite{gsm8k}) to evaluate generalization abilities. All evaluation and benchmark details can be found in Appendix §\ref{sec:benchmark_details}.

\noindent\textbf{{Baselines.}}
We compare with the following competitive baselines.
Closed-Source Models: GPT-3.5-Turbo \cite{chatgpt} and GPT-4-Turbo \cite{GPT-4}. Open-Source Models: Llama 2 \cite{llama2}, CodeLlama \cite{codellama}, StarCoder \cite{starcoder}, WizardCoder \cite{wizardcoder}, DeepSeek-Coder \cite{deepseek-coder}, and Magicoder \cite{wei2023magicoder}.

\noindent\textbf{{Supervised fine-tuning.}}
We adopt Llama-2-7B, CodeLlama-Python-7B, and DeepSeek-Coder-Base-6.7B as the base models and fine-tune all the base models for 2 epochs using 32 NVIDIA A100-80GB GPUs. We set the initial learning rate at 1e-4. We use Adam optimizer \cite{adam} and choose a batch size of 2 with a sequence length of 8192.

\definecolor{highlightcolor}{RGB}{233,247,217}
\begin{table*}[t]
\centering
\caption{Results of pass@1 on HumanEval (HumanEval+) and MBPP (MBPP+) benchmarks. We report the results of HumanEval and MBPP consistently from the EvalPlus \cite{humaneval+} and the \textbf{bold} scores denote the best performance among models of the same size.}
% \vspace{-3pt}
% \footnotesize
\footnotesize
% \fontsize{9pt}{5pt}\selectfont % 设置字体大小和行间距
\setlength\tabcolsep{2.7pt} 
\begin{tabular}{llcccc}
\toprule
\multicolumn{1}{l|}{\textbf{Model}}             & \textbf{Params}  & \textbf{Base Model} & \textbf{HumanEval (+)} & \textbf{MBPP (+)} & \textbf{Average (+)} \\ \midrule
\multicolumn{6}{l}{\textit{\textbf{Closed-source Models}}}                                                                                                     \\ \midrule
\multicolumn{1}{l|}{GPT-3.5-Turbo \cite{chatgpt}}              & -                      & -                   & {72.6} ({65.9})               & {81.7} ({69.4})         & {77.2} ({67.7})             \\
\multicolumn{1}{l|}{GPT-4-Turbo \cite{GPT-4}}                & -                      & \textbf{-}          & \textbf{85.4 (81.7)}      & \textbf{83.0 (70.7)} & \textbf{84.2 (76.2)}    \\ \midrule
\multicolumn{6}{l}{\textit{\textbf{Open-source Models}}}                                                                                                      \\ \midrule
\multicolumn{1}{l|}{Llama 2-Chat \cite{llama2}}                & 70B                  & Llama 2              & 31.7 (26.2)               & 52.1 (38.6)         & 41.9 (32.4)            \\
% \multicolumn{1}{l|}{Lemur-Chat (\cite{xu2023lemur})}                 & 70B                    & Lemur               & 61.0 ()              & 55.5 ()         & 58.3 ()            \\ 
\multicolumn{1}{l|}{CodeLlama-Python \cite{codellama}}           & 70B                 & Llama 2                   &  57.9 (50.0)              &  72.4 (52.4)          &  65.2 (51.2)            \\
\multicolumn{1}{l|}{CodeLlama-Instruct \cite{codellama}}           & 70B                 & CodeLlama                   & \textbf{65.2 (58.5)}              & \textbf{73.5 (55.1)}          & \textbf{69.4 (56.8)}            \\ \grayrule
% \multicolumn{1}{l|}{CodeLlama (\cite{codellama})}                  & 34B                   & Llama 2                   & 48.8 ()              & 55.0 ()         & 51.9 ()            \\
\multicolumn{1}{l|}{CodeLlama-Python \cite{codellama}}           & 34B                 & Llama 2                   & 51.8 (43.9)              & 67.2 (50.4)          & 59.5 (47.2)            \\
\multicolumn{1}{l|}{WizardCoder-CL \cite{wizardcoder}}         & 34B                   & CodeLlama-Python           & 73.2 (56.7)      & 73.2 (51.9)          & 73.2 (54.3)            \\ 
\multicolumn{1}{l|}{DeepSeek-Coder-Instruct \cite{deepseek-coder}}         & 33B                   & DeepSeek-Coder-Base           & \textbf{78.7} (\textbf{67.7})      &  \textbf{78.7} (\textbf{59.7})          &  \textbf{78.7} (\textbf{63.7})            \\ \grayrule
% \multicolumn{1}{l|}{CodeT5+ (\cite{codeT5})}           & 16B                     & -                   & 31.7 (26.2)              & 54.6 (44.4)         & 48.0 (39.8)            \\
% \multicolumn{1}{l|}{CodeGen (\cite{nijkamp2022codegen})}           & 16B                     & -                   & 32.9 (27.4)              & 52.6 (43.6)         & 48.0 (39.8)            \\
\multicolumn{1}{l|}{StarCoder \cite{starcoder}}                  & 15B                    & -                   & 34.1 (33.5)               & 55.1 (43.4)          & 44.6 (38.5)            \\
% \multicolumn{1}{l|}{CodeLlama (\cite{codellama})}                  & 13B                    & Llama 2                   & 36.0 ()              & 47.0 ()         & 41.5 ()            \\
\multicolumn{1}{l|}{CodeLlama-Python \cite{codellama}}           & 13B                    & Llama 2                   & 42.7 (36.6)              & 61.2 (\textbf{45.6})         & 52.0 (41.1)            \\
\multicolumn{1}{l|}{WizardCoder-SC \cite{wizardcoder}}                & 15B                    & StarCoder           & \textbf{51.9} (\textbf{45.7})              & \textbf{61.9} (44.9)          & \textbf{56.9} (\textbf{45.3})            \\ \grayrule
\multicolumn{1}{l|}{Llama 2 \cite{llama2}}                     & 7B                    & -                   & 14.0 (10.4)              & 26.1 (17.5)         & 20.1 (14.0)            \\
% \multicolumn{1}{l|}{CodeLlama (\cite{codellama})}                  & 7B                     & Llama 2                   & 33.5 ()              & 41.4 ()         & 37.5 ()            \\
\multicolumn{1}{l|}{StarCoder \cite{starcoder}}                  & 7B                    & -                   & 24.4 (21.3)               & 33.1 (29.2)          & 28.8 (25.3)            \\
% \multicolumn{1}{l|}{CodeT5+ (\cite{codeT5})}           & 6B                     & -                   & 29.3 (23.8)              & 51.9 (40.9)         & 48.0 (39.8)            \\
% \multicolumn{1}{l|}{CodeGen (\cite{nijkamp2022codegen})}           & 6B                     & -                   & 29.3 (25.6)              & 49.9 (42.1)         & 48.0 (39.8)            \\
\multicolumn{1}{l|}{CodeLlama-Python \cite{codellama}}           & 7B                     & Llama 2                   & 37.8 (33.5)              & 57.6 (42.4)         & 47.7 (38.0)            \\
\multicolumn{1}{l|}{WizardCoder-CL \cite{wizardcoder}}         & 7B                   & CodeLlama-Python           & 48.2 (42.1)      & 56.6 (42.4)          & 52.4 (42.3)            \\
% \multicolumn{1}{l|}{WaveCoder-CL (\cite{wavecoder})}               & 7B                    & CodeLlama           & 48.1 ()              & 47.2 ()         & 47.7 ()            \\
\multicolumn{1}{l|}{DeepSeek-Coder-Base \cite{deepseek-coder}}        & 6.7B                  & -                   & 47.6 (41.5)    & 70.2 (53.6) & 58.9 (47.6)            \\
\multicolumn{1}{l|}{Magicoder-CL \cite{wei2023magicoder}}               & 7B                     & CodeLlama-Python    & 60.4 (49.4)              & 64.2 (46.1)         & 62.3 (47.8)            \\
% \multicolumn{1}{l|}{WaveCoder-DS (\cite{wavecoder})}               & 6.7B                   & DeepSeek-Coder-Base & 62.8 ()              & 62.4 ()         & 62.6 ()            \\
\multicolumn{1}{l|}{MagicoderS-CL \cite{wei2023magicoder}}              & 7B                     & CodeLlama-Python    & 70.7 (60.4)              & 68.4 (49.1)   & 69.6 (54.8)   \\
\multicolumn{1}{l|}{Magicoder-DS \cite{wei2023magicoder}}              & 6.7B                     & DeepSeek-Coder-Base    & 66.5 (55.5)              & 75.4 (55.6)   & 71.0 (55.6)   \\
\multicolumn{1}{l|}{DeepSeek-Coder-Instruct \cite{deepseek-coder}}         & 6.7B                   & DeepSeek-Coder-Base           & 73.8 (69.5)      &  72.7 (55.6)          & 73.3 (62.6)            \\
\multicolumn{1}{l|}{MagicoderS-DS \cite{wei2023magicoder}}              & 6.7B                     & DeepSeek-Coder-Base    & 76.8 (65.2)              & 75.7 (56.1)   & 76.3 (60.7)   \\ 
\rowcolor{highlightcolor} \multicolumn{1}{l|}{\textit{\textbf{AlchemistCoder-L (ours)}}}  & 7B                    & Llama 2              & 56.7 (52.4)             & 54.5 (49.6)         & 55.6 (51.0)            \\
\rowcolor{highlightcolor}\multicolumn{1}{l|}{\textit{\textbf{AlchemistCoder-CL (ours)}}} & 7B                     & CodeLlama-Python    & {74.4} ({68.3})               & {68.5} (55.1)          & {71.5} (61.7)             \\
\rowcolor{highlightcolor}\multicolumn{1}{l|}{\textit{\textbf{AlchemistCoder-DS (ours)}}} & 6.7B                & DeepSeek-Coder-Base &  \textbf{79.9} (\textbf{75.6})                  &       \textbf{77.0} (\textbf{60.2})        &   \textbf{78.5} (\textbf{67.9})               \\ \bottomrule
\end{tabular}
\label{table: humaneval and mbpp}
\vspace{-7pt}
\end{table*}

\subsection{Evaluation on code generation task}
\noindent\textbf{Results on python code generation.}
We first access HumanEval and MBPP to evaluate the capability of the \textit{AlchemistCoder} series for Python code generation. 
% As two prominent benchmarks for code generation tasks, HumanEval and MBPP comprise a substantial collection of hand-written Python programming tasks with test cases. 
These benchmarks necessitate models to generate code based on the function definitions and subsequently pass the test cases. Models are evaluated in zero-shot on HumanEval and 3-shot on MBPP. The comprehensive comparisons in Tab. \ref{table: humaneval and mbpp} and Fig. \ref{figure: Comparison on HumanEval and MBPP} demonstrate the impressive capabilities of \textit{AlchemistCoder} models. From the results, \textit{AlchemistCoder-L} attains a remarkable performance boost of 42.7\% and 28.4\% pass@1 scores on HumanEval and MBPP respectively, compared to Llama 2-7B. Notably, \textit{AlchemistCoder-DS} elevates the pass@1 scores to 79.9\% and 77.0\% on these benchmarks, holding an overall improvement of 33.3\%. Moreover, our \textit{AlchemistCoder} series with 7B parameters outperforms larger models (\textit{e.g.}, WizardCoder-CL-34B and CodeLlama-Instruct-70B) and rivals with GPT-3.5-Turbo, significantly bridging the performance gap between closed-source and open-source models.

\begin{table}[t]
    \centering
    \begin{minipage}{0.46\textwidth}
        \caption{Results of pass@1 on HumanEval-X. We present the multilingual code capabilities of our \textit{AlchemistCoder} with the respective base models and competitors (6.7B/7B).}
        \vspace{-6pt}
        \centering
        \scriptsize
        \setlength\tabcolsep{2.3pt} 
        \begin{tabular}{l|cccccc}
        \toprule
        \textbf{Model}    &  \textbf{Python} & \textbf{C++}  & \textbf{Go}   & \textbf{Java} & \textbf{JS} & \textbf{Avg} \\ \midrule
        Llama 2                        & 14.0            & 11.0          & 6.1           & 11.0           & 14.0                & 11.2              \\
        CodeLlama                        & 31.7            & 27.4          & 12.8           & 25.6           & 32.9                & 26.1              \\
        \rowcolor{highlightcolor}\textit{\textbf{AlchemistCoder-L}}                & \textbf{56.7}            & \textbf{31.1}          & \textbf{25.6}          & \textbf{45.1}          & \textbf{41.5}                & \textbf{37.1}             \\ \grayrule
        CodeLlama-Python               & 37.8            & 33.5          & 30.5          & 39.6          & 35.4                & 35.4             \\
        MagicoderS-CL               & 68.3            & 47.6          & 39.6          & 34.8          & \textbf{57.9}                & 49.6             \\
        \rowcolor{highlightcolor}\textit{\textbf{AlchemistCoder-CL}}              & \textbf{74.4}   & \textbf{53.1} & \textbf{42.7} & \textbf{64.0} & {52.4}       & \textbf{57.3}    \\ \grayrule
        DeepSeek-Coder-Base               & 47.6            & 45.1          & 38.4          & 56.1          & 43.9                & 46.2             \\
        MagicoderS-DS               & 72.6            & \textbf{63.4}          & 51.8          & 70.7          & 66.5                & 65.0             \\
        \rowcolor{highlightcolor}\textit{\textbf{AlchemistCoder-DS}}              & \textbf{79.9}   & {62.2} & \textbf{59.8} & \textbf{72.0} & \textbf{68.9}       & \textbf{68.6}    \\ \bottomrule
        \end{tabular}
        \label{table: humanevalx}
    \end{minipage}\hfill
    \begin{minipage}{0.52\textwidth}
        \caption{Pass@1 results of models with 6.7B/7B parameters on DS-1000. pd, np, tf, sp, skl, torch, and plt represent Pandas, Numpy, Tensorflow, Scipy, Sklearn, Pytorch, and Matplotlib, respectively.}
        \vspace{-6pt}
        \centering
        \scriptsize
        \setlength\tabcolsep{2.3pt} 
        \begin{tabular}{l|cccccccc}
        \toprule
        \textbf{Model}               & \textbf{pd}   & \textbf{np}   & \textbf{tf}    & \textbf{sp}   & \textbf{skl}  & \textbf{torch} & \textbf{plt}  & \textbf{All} \\ \midrule
        Llama 2             & 2.4  & 7.3  & 6.7   & 6.6  & 2.6  & 1.5   & 7.7  & 5.0     \\
        CodeLlama             & 12.0  & \textbf{27.7}  & 17.8   & \textbf{13.2}  & 12.2  & \textbf{20.6}   & 15.5  & 17.0     \\
        \rowcolor{highlightcolor}\textit{\textbf{AlchemistCoder-L}}    & \textbf{13.4} & 22.7  & \textbf{31.1}   & 11.3  & \textbf{25.2} & 8.8   & \textbf{29.0} & \textbf{20.2}     \\ \grayrule 
        CodeLlama-Python             & 16.2  & 16.4  & 15.6   & 17.9  & 20.0  & 22.1   & 38.7  & 21.0     \\
        % Magicoder-CL             & 0.3  & 0.9  & 0.0   & 0.0  & 0.9  & 0.0   & 0.0  & 0.3     \\
        MagicoderS-CL             & 25.1  & 40.9  & 35.6   & 29.3  & 36.5  & 38.2   & 51.0  & 36.7     \\
        \rowcolor{highlightcolor}\textit{\textbf{AlchemistCoder-CL}}    & \textbf{30.9} & \textbf{43.6}  & \textbf{46.7}   & \textbf{30.2}  & \textbf{37.4} & \textbf{41.2}   & \textbf{52.3} & \textbf{40.3}     \\ \grayrule
        DeepSeek-Coder-Base & 21.3 & 35.0 & 26.7 & 23.6 & 34.8 & 25.0  & 34.8 & 28.7    \\
        MagicoderS-DS             & 30.6  & 46.8  & 44.2   & 30.2  & 33.0  & 29.7   & 45.2  & 37.1     \\
        \rowcolor{highlightcolor}\textit{\textbf{AlchemistCoder-DS}}    & \textbf{32.0} & \textbf{51.7}  & \textbf{44.5}   & \textbf{33.1}  & \textbf{38.4} & \textbf{33.8}  & \textbf{49.8} & \textbf{40.5}    \\ 
        \bottomrule
        \end{tabular}
        \label{table: DS-1000}
    \end{minipage}
\vspace{-5pt}
\end{table}

\noindent\textbf{Results on multilingual code generation.}
We compare the pass@1 accuracy of the base models and the corresponding fine-tuned \textit{AlchemistCoder} models on Humaneval-X \cite{humanevalx}. The results presented in Tab. \ref{table: humanevalx} demonstrate that the \textit{AlchemistCoder} series exhibits great improvements (exceeding 50\%) for multilingual code generation, delivering comprehensive code capabilities.

\noindent\textbf{Results on code generation for data science.}
We further conduct the evaluation of data science code completion on DS-1000 \cite{ds-1000}. According to Tab. \ref{table: DS-1000}, \textit{AlchemistCoder} models exhibit a notable improvement of up to 19.2\% in overall performance compared to the base models. Particularly, \textit{AlchemistCoder-CL} achieves an astonishing overall accuracy of 40.3\% with relatively better performance in all libraries, demonstrating powerful capabilities in data science workflows.

\begin{figure}[h]
    \centering
    \begin{minipage}{0.47\textwidth}
        \centering
        \includegraphics[width=1.00\textwidth]{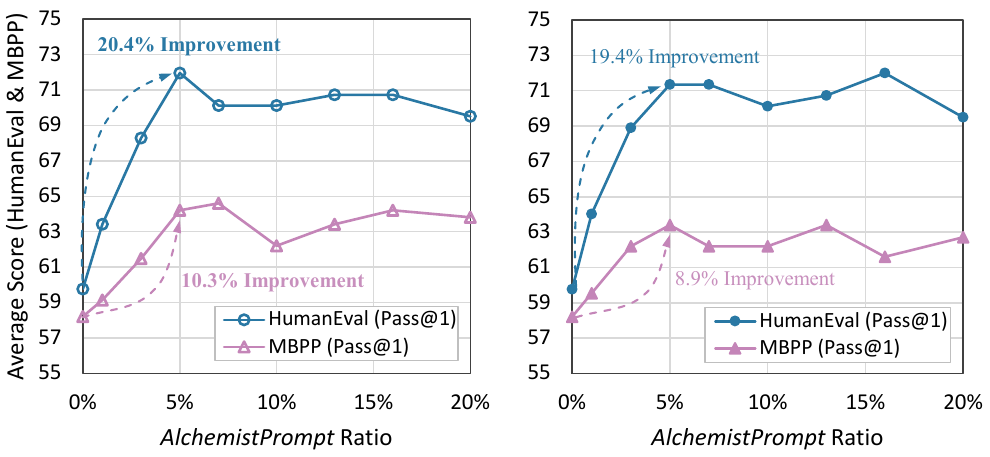}
        \vspace{-17pt}
        \figcaption{Ablation study on the proportion of \promptname{}-customized data conducted on \textit{AlchemistCoder-CL-7B}. Left: Augment the original data. Right: Replace the original data.}
        \label{figure: Ablation of ALCMEMIST-PROMPT Ratio}
    \end{minipage}\hfill
    \begin{minipage}{0.51\textwidth}
        \tabcaption{Ablation study on the effectiveness of \promptname{}s and code understanding tasks for the \textit{AlchemistCoder-CL-7B} model, evaluated on the HumanEval and MBPP benchmarks.}
        \vspace{3pt}
        \centering
        {\scriptsize
        \setlength\tabcolsep{2.1pt} 
        \begin{tabular}{cccc|cc}
        \toprule
        \textbf{\makecell{Alchemist\\Prompt}} & \textbf{\makecell{Instruction\\Evolution}} & \textbf{\makecell{Data\\Filtering}} & \textbf{\makecell{Code\\Review}} & \textbf{\makecell{HumanEval\\(Pass@1)}}     & \textbf{\makecell{MBPP\\(Pass@1)}}            \\ \midrule
        -                     &  -                     & -              & -           & 59.8            & 58.2              \\ 
        \ding{51}                     &  -                     & -              & -           & 72.0            & 63.4              \\ 
        \ding{51}                     &  \ding{51}                     & -              & -           & 71.3          & 65.8                   \\
        \ding{51}                     &  \ding{51}                                       &  \ding{51}              & -           & 73.8          & 67.7                \\
        -                     &  \ding{51}                                       &  \ding{51}              & \ding{51}           & 65.2          & 64.6                \\
        \rowcolor{highlightcolor}\ding{51}                     & \ding{51}                     & \ding{51}              & \ding{51}           & \textbf{74.4} & \textbf{68.5}  \\ \bottomrule
        \end{tabular}}
        \label{table: Ablation of Code Comprehension Task}
    \end{minipage}
\end{figure}

\subsection{Ablation study}
\noindent\textbf{The Recipe of \promptname{}s.}
As illustrated in Sec. \ref{section: AlchemistPrompt}, \promptname{}s can further align the instructions and responses of data samples and harmonize the domain gap between multiple sources. To find the appropriate recipe of \promptname{}s that maintains a balance between data diversity and domain gap, we conduct ablation experiments on the proportion (0\% to 20\%) of data customized by \promptname{}s. We adopt two settings: (a) augment the original data with its customized variant and report the results of fine-tuning for 2 epochs on CodeLlama-Python-7B; (b) replace the original data and report the results of fine-tuning for the same steps (\textit{i.e.}, keeping the number of tokens used consistent). 
As shown in Fig. \ref{figure: Ablation of ALCMEMIST-PROMPT Ratio}, \textit{AlchemistCoder} is particularly enhanced when the proportion of customized data increases from 1\% to 5\%, and nearly peaks in performance at 5\%. Thus, we introduce \promptname{}s into 5\% of the training set to balance the performance gain and the generation cost. 
Additionally, both two strategies effectively enhance the performance and validate the efficacy of our approach. To push the limit of \textit{AlchemistCoder}, we employ the augmentation strategy in our performance experiments.

\noindent\textbf{Efficacy of the code comprehension tasks.}
We conduct an ablation study on the key components of the code comprehension tasks to ascertain their individual contributions to the overall performance. As reported in Tab. \ref{table: Ablation of Code Comprehension Task}, compared to the baselines (the first and second rows), the model demonstrates enhanced performance on both benchmarks following the incremental incorporation of code comprehension task data. Notably, the improvement (5.1\% regard to the pass@1 metric) is particularly remarkable on MBPP. This indicates the significant contribution of all code comprehension tasks to furthering programming capabilities.

\subsection{Analytical study}
% \noindent\textbf{Data Distribution Analysis.}
\begin{figure}[t]
\centering
\includegraphics[width=0.95\textwidth]{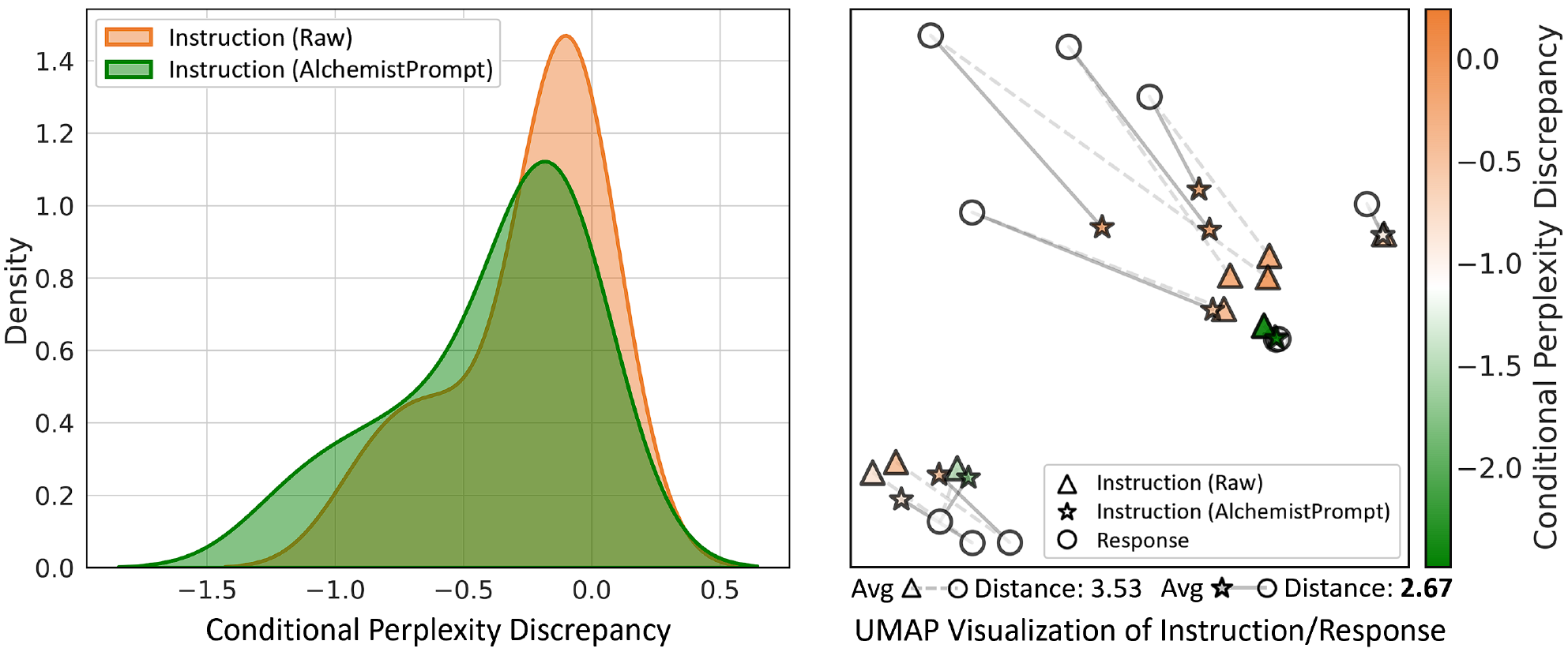}
\vspace{-3pt}
\caption{In-depth analysis of the efficacy from \promptname{}s. Left: Kernel Density Estimation of conditional perplexity discrepancy. Right: UMAP visualization of 10 instruction/response groups. }
\label{figure: Perplexity and UMAP}
\vspace{-7pt}
\end{figure}

\noindent\textbf{\promptname{}s harmonize the discrepancy between instructions and responses.}
To in-depth verify the efficacy of \promptname{}s, we calculate the perplexities of the model in generating responses under given conditions (instructions), \textit{i.e.}, the difference between $\mathrm{Perplexity(conditional\_instruction + response)}$ and $\mathrm{Perplexity(response)}$, called Conditional Perplexity Discrepancy (CPD). Specifically, we adopt the instructions before and after customization by \promptname{}s for comparison, and provide the Kernel Density Estimation of CPD in Fig. \ref{figure: Perplexity and UMAP}. Clearly, the latter (green) gains smaller overall CPD values, indicating that \promptname{}s are beneficial for prediction and can provide effective contextual information. Furthermore, we randomly select 10 groups of these samples and use UMAP \cite{mcinnes2018umap} to map their feature representations into a 2-D space in the right of Fig. \ref{figure: Perplexity and UMAP}. From the fact that the solid lines are generally shorter than the dashed lines, our \promptname{}s can harmonize the discrepancy between instructions and responses, leading to higher-quality data for attaining improved instruction-following ability.

\begin{figure}[t]
    \centering
    \begin{minipage}{0.49\textwidth}
        \tabcaption{Results of models (6.7B/7B) on various benchmarks, including MMLU for multitask language understanding, BBH for comprehensive reasoning, and GSM8K for mathematical ability.}
        % \vspace{-6pt}
        \centering
        \scriptsize
        \setlength\tabcolsep{4pt} 
        \begin{tabular}{l|cccc}
        \toprule
        \textbf{Model}     & \textbf{MMLU} & \textbf{BBH}  & \textbf{GSM8K} & \textbf{Avg}  \\ \midrule
        Llama 2                        & 41.1          & 34.6          & 16.8           & 30.8          \\
        CodeLlama                         & 31.5          & \textbf{42.7}          & 14.4           & 29.5          \\
        \rowcolor{highlightcolor}\textit{\textbf{AlchemistCoder-L}}                & \textbf{43.9} & \textbf{42.7} & \textbf{25.0}  & \textbf{37.2} \\ \grayrule
        CodeLlama-Python               & 26.1          & 26.7          & 6.6            & 19.8          \\
        MagicoderS-CL                & 33.0          & \textbf{41.5}          & 18.8            & 31.1          \\
        \rowcolor{highlightcolor}\textit{\textbf{AlchemistCoder-CL}}               & \textbf{42.1}          & 39.3          & \textbf{20.2}           & \textbf{33.9}          \\ \grayrule
        DeepSeek-Coder-Base               & 34.0          & 12.8          & 22.0            & 22.9          \\
        MagicoderS-DS               & 34.4          & {43.8}          & 14.3            & 30.8          \\
        \rowcolor{highlightcolor}\textit{\textbf{AlchemistCoder-DS}}               & \textbf{38.5}          & \textbf{45.6}          & \textbf{28.0}           & \textbf{37.4}          \\ \bottomrule
        \end{tabular}
        \label{table: Reasoning and Math Benchmarks}
    \end{minipage}\hfill
    \begin{minipage}{0.49\textwidth}
        \centering
        \includegraphics[width=0.97\textwidth]{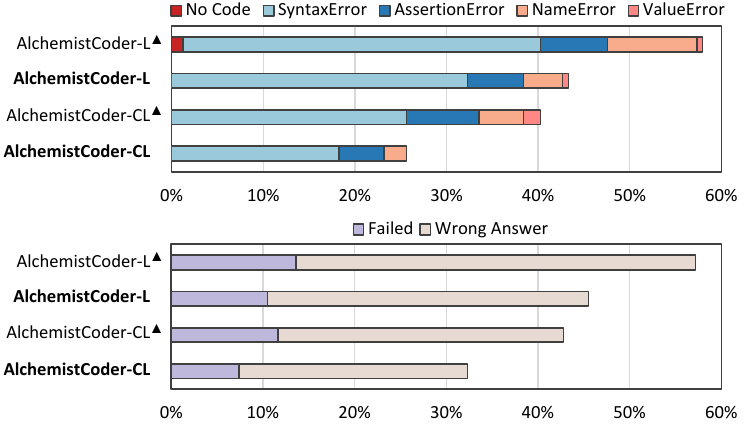}
        \vspace{-7pt}
        \figcaption{Analysis of error case proportions on HumanEval (top) and MBPP (bottom). $^{\blacktriangle}$ represents the models fine-tuned without \promptname{}s and the code comprehension task data.}
        \label{figure: Error Cases Analysis}
    \end{minipage}
\vspace{-7pt}
\end{figure}

\noindent\textbf{\textit{AlchemistCoder} models are better generalists.}
To further analyze the comprehensive capabilities of our \textit{AlchemistCoder}, we conduct evaluations on more diversified benchmarks, including MMLU \cite{mmlu} for multitask language understanding, BBH \cite{bbh} for comprehensive reasoning, and GSM8K \cite{gsm8k} for mathematical ability. The results are presented in Tab. \ref{table: Reasoning and Math Benchmarks} and illustrate that the \textit{AlchemistCoder} models exhibit an overall performance increase of 6.4\%, 13.6\%, and 14.5\% over the base models Llama 2, CodeLlama-Python, and DeepSeek-Coder-Base, respectively. Notably, CodeLlama-Python presents inferior performance on these benchmarks relative to Llama 2, indicating the discrepancy between natural language processing and code capabilities of open-source models. Such divergence can be attributed to ``catastrophic forgetting'' \cite{catastrophic_forgetting_ref_1, catastrophic_forgetting_ref_2, catastrophic_forgetting_ref_3}, often occurring when fine-tuning is exclusively concentrated on domain-specific data. By leveraging harmonized multi-source data, our \textit{AlchemistCoder} series achieves more multifaceted and comprehensive capabilities.

\noindent\textbf{Error case analysis.}
To meticulously dissect the improvements brought by our method, we provide an analysis of error cases on HumanEval and MBPP. We compare models before and after the introduction of \promptname{}s and code understanding task data. The bar chart shown in Fig. \ref{figure: Error Cases Analysis} (top) indicates that these two types of key data help to better handle compilation errors (\textit{i.e.}, SyntaxError, NameError, and ValueError), and eliminate the occurrence of no code written in the responses. On the other hand, the results of Fig. \ref{figure: Error Cases Analysis} (bottom) on MBPP suggest that the \textit{AlchemistCoder} series incorporated with these two types of data attains stronger programming logic, as evidenced by the clear reduction in the `Wrong Answer' error cases.

\section{Related Work}
\noindent\textbf{Code large language models.}
% Code LLMs, pretrained and fine-tuned on code data, have demonstrated remarkable capabilities in tasks like code generation and code parsing. 
Early researches \cite{humaneval,nijkamp2022codegen, starcoder} focus on collecting massive amounts of code data to develop pretrained Code LLMs. Recent efforts \cite{wizardcoder, wavecoder, wei2023magicoder} are dedicated to fine-tuning these pretrained models with specific instructional data to further the coding abilities. For instance, WizardCoder \cite{wizardcoder} and Magicoder \cite{wei2023magicoder} construct their instruction tuning datasets based on CodeAlpaca \cite{codealpaca} and the stack \cite{kocetkov2022stack} dataset, respectively. In this work, we develop the \textit{AlchemistCoder} series by instruction tuning on optimized multi-source data instead of single-category data as in previous methods, endowing astonishing and harmonized code capability.

\noindent\textbf{Instruction tuning.}
Instruction tuning aims to enhance LLMs via fine-tuning pre-trained LLMs using samples of instruction/response pairs. Obtaining high-quality data for instruction tuning is typically challenging and extensive works have been dedicated to this endeavor. For instance, Alpaca \cite{alpaca} employs self-instruct \cite{selfinstruct} to generate instruction-following demonstrations. 
% Following Alpaca, Peng \textit{et al.} \cite{peng2023instruction} adopt GPT-4 as a distillation teacher. 
WizardLM  \cite{xu2023wizardlm} introduces Evol-Instruct and transforms the instruction data into more complex variants. 
In addition to Evol-Instruct, we also incorporate the data construction process itself as a form of data into the training. Moreover, although previous works \cite{ivison2023camels, tulu, wang2023openchat} utilize multiple fine-tuning datasets, we harmonize multi-source data at a fine-grained level.
% Unlike previous researches that process training data uniformly, we customarily polish the quality of instructions and strive to maximize the use of multi-source data for code generation.

\noindent\textbf{Learning from hindsight.}
The concept of learning from hindsight \cite{chain_of_hindsight} has been explored in goal-conditioned learning \cite{kaelbling1993learning, ganguli2022red}. Andrychowicz \textit{et al.} \cite{hindsight_openai_2017} introduce Hindsight Experience Replay (HER) to re-label rewards and facilitate learning from sparse feedback retrospectively. Korbak \textit{et al.} \cite{korbak2023pretraining} study the influence of human preferences during pre-training, showing improved performance when models are aligned with human preferences. 
Previous work primarily serves as an alternative to RLFT, utilizing HER to leverage (suboptimal) historical data for model learning. We focus on constructing multi-source data and harmonizing the inherent conflicts within multi-source data through hindsight, to fully tap into the potential of base models.
% In this paper, we present the \promptname{} with the key idea of applying hindsight to polish instructions. 

% Our work focuses on fine-tuning pre-trained language models and explores the positive performance benefits from aligning with the ``preferences'' from \promptname{}s generated by language models.
\section{Conclusion}
In this paper, we propose an effective framework for integrating multi-source data to fine-tune Code LLMs, addressing the limitations in quality and diversity inherent within a single-source dataset. This is a non-trivial paradigm and we pioneer to unveil inherent conflicts in multi-source code corpora. To resolve this challenge, we innovatively design data-specific \promptname{}s, inspired by hindsight relabeling. Additionally, we make the first effort of integrating the data construction process as code comprehension tasks into the training process. These key concepts enhance the diversity, quality, and cost-effectiveness of code fine-tuning data, facilitating the development of the \textit{AlchemistCoder} series models with significantly improved and comprehensive coding capabilities.

% In this paper, we develop {\textit{AlchemistCoder}}, a series of enhanced Code LLMs fine-tuned on multi-source data. To achieve this, we introduce data-specific \promptname{}s leveraging the idea of hindsight to harmonize multi-source data, and design code comprehension tasks to provide data with more dimensions. Performance experiments verify the harmonized and superior code capabilities of the \textit{AlchemistCoder} models. Additionally, extensive analytical studies have been well-designed and delved deeply into the efficacy of our method.

{
\small
\bibliographystyle{plainnat}
\bibliography{custom}
}

\section*{Appendix}
\appendix
% \onecolumn
In the Appendix sections, we discuss limitations (§\ref{limitations}), ethical considerations and broader impacts (§\ref{Ethical Considerations and Broader Impacts}), benchmark and evaluation details (§\ref{sec:benchmark_details}), and more details of our \textit{AlchemistCoder} fine-tuning data (§\ref{AlchemistCoder Dataset Details}).

\setcounter{table}{0} 
\setcounter{figure}{0}
\setcounter{equation}{0}
\renewcommand{\thetable}{A\arabic{table}}
\renewcommand\thefigure{A\arabic{figure}} 
\renewcommand\theequation{A\arabic{equation}}

\section{Limitations}
\label{limitations}
% In this work, we introduce the \promptname{} integrating the idea of hindsight. 
Currently, GPT-4 holds an advantage in generating high-quality responses, and thus has been chosen as our \textit{Alchemist} model. 
Compared to methods that heavily rely on GPT-4 to generate entire new datasets, we have been striving to minimize our dependence on GPT-4. Instead of using GPT-4 to generate data from scratch, we optimize a small amount of data. In the ablation experiments shown in Figure 7, we have verified that achieving optimal performance only requires using GPT-4 to generate \promptname{}s for 5\% of the data. Furthermore, the generation tasks we designed only require very short responses (less than 50 words), significantly reducing the token usage of GPT-4.
Despite these efforts, the generation of \promptname{}s is still a significant cost. We will explore fine-tuning open-source models to achieve the free generation of \promptname{}s in the future.

\section{Ethical Considerations and Broader Impacts}
\label{Ethical Considerations and Broader Impacts}
We use publicly available datasets, benchmarks, and models for training and evaluation, free from any possible harm toward individuals or groups. The generated data from LLMs are relevant to code-related tasks and no personal identification information is involved. Furthermore, we adopt ChatGPT to polish the writing and assist with language. For broader impacts, \textit{AlchemistCoder} enhances code generation and generalization through multi-source fine-tuning, promising improved software development efficiency, democratization of programming, and educational benefits. However, it also raises concerns about malicious use, intellectual property issues, and skill degradation. To ensure the responsible release of \textit{AlchemistCoder} models, we will implement controlled access, provided usage guidelines, and engaged with the research community, thereby mitigating the risks of misuse or dual-use.

\section{Benchmark and Evaluation Details}\label{sec:benchmark_details}

\begin{wrapfigure}{r}{0.4\textwidth}
\centering
\vspace{-7pt}
\includegraphics[width=0.38\textwidth]{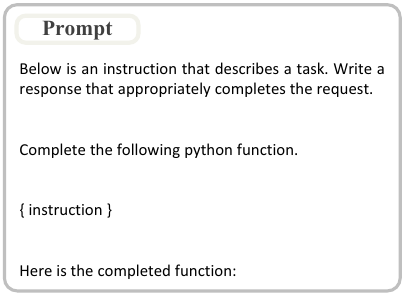} % Reduce the figure size so that it is slightly narrower than the column.
\vspace{-7pt}
\caption{Prompt used to evaluate on HumanEval and HumanEval+.}
\label{figure: HumanEval Evaluation Prompt}
\end{wrapfigure}

\subsection{HumanEval/HumanEval+}
HumanEval \cite{humaneval} and HumanEval+ \cite{humaneval+} are benchmarks for assessing LLMs' code generation, focusing on functional correctness. HumanEval+ expands on HumanEval by significantly increasing test cases through EvalPlus, using LLM and mutation strategies for more rigorous evaluation. This approach reveals performance drops in models like GPT-4 and ChatGPT against challenging tests, emphasizing the need for diverse test scenarios to accurately evaluate LLMs' coding abilities. For evaluation on HumanEval and HumanEval+, we adopt the prompt designed for HumanEval/HumanEval+ tasks shown in Fig. \ref{figure: HumanEval Evaluation Prompt}. 
Following prior works \cite{humanevalx, chen2023teaching, wei2023magicoder}, we use the greedy decoding strategy and focus on comparing the pass@1 metric. 
% We generate 50 samples to estimate the scores with $\mathrm{temperate=1.0}$ and $\mathrm{top}\mathrm{\_}\mathrm{p=1.0}$. Additionally, we maintain this setup in all other benchmarks.

\subsection{MBPP/MBPP+}
The MBPP (Mostly Basic Python Programming) benchmark \cite{mbpp} consists of around 1,000 Python challenges, crowd-sourced to test basic programming skills, including fundamentals and standard library use. Aimed at beginners, each challenge offers a description, solution, and three tests for verifying solution accuracy. MBPP+ \cite{humaneval+} is an extension of the MBPP benchmark, utilizing a subset of hand-verified problems from MBPP-sanitized to ensure tasks are well-defined and unambiguous. For the evaluation on MBPP and MBPP+, we adopt the three-shot prompt shown in Fig. \ref{figure: MBPP Evaluation Prompt}. 

\subsection{HumanEval-X}
HumanEval-X \cite{humanevalx} is a comprehensive benchmark that assesses the capabilities of code generation models across multiple programming languages, including Python, C++, Java, JavaScript, and Go. It consists of 820 meticulously created data samples, each accompanied by test cases, making it an invaluable resource for evaluating and improving multilingual code generation models. The benchmark aims to provide insights into the models' proficiency in solving diverse coding challenges and their accuracy in generating functionally correct code in different languages. For evaluation on HumanEval-X, we do not use specific prompts and follow the original test prompts.

\subsection{DS-1000} 
The DS-1000 benchmark \cite{ds-1000} adapts 1000 different data science coding problems each with unit tests from StackOverflow and checks both execution semantics and surface-form constraints. These realistic problems are drawn from seven popular data science libraries in Python, including Matplotlib (plt), NumPy (np), Pandas (pd), SciPy (scp), Scikit-Learn (sk), PyTorch (py), and TensorFlow (tf). DS-1000 has two modes: completion and insertion, and here we only evaluate completion, as the basic CodeLlama-Python does not support insertion. For evaluation on DS-1000, we do not use specific prompts and follow the original test prompts.

\subsection{MMLU}
The Massive Multitask Language Understanding (MMLU) benchmark \cite{mmlu} is an evaluation framework designed to measure the depth and breadth of knowledge that LLMs possess. It accomplishes this by testing these models across 57 varied tasks in both zero-shot and few-shot scenarios. The tasks encompass a wide array of topics, including basic math, American history, computer science, law, and more, challenging the models to leverage their acquired knowledge to solve complex problems. MMLU seeks to emulate the multifaceted way in which human knowledge and problem-solving skills are assessed, offering a comprehensive gauge of a model's ability to understand and apply information across multiple domains. For evaluation on MMLU, we do not use specific prompts and follow the original test prompts.

\subsection{BBH}
The BIG-Bench Hard (BBH) Benchmark \cite{bbh} is a specialized evaluation framework tailored to rigorously test the capabilities of LLMs. This benchmark targets a selection of tasks that have historically proven challenging for LLMs, focusing on areas where models typically do not exceed average human performance. The BBH Benchmark aims to push the boundaries of what LLMs can achieve by emphasizing complex reasoning, deep understanding, and nuanced interpretation, setting a high bar for model development and performance evaluation. For evaluation on BBH, we do not use specific prompts and follow the original test prompts.

\subsection{GSM8K}
The GSM8K (Grade School Math 8,000) benchmark \cite{gsm8k} serves as a rigorous evaluation framework for testing the mathematical problem-solving prowess of LLMs. This benchmark comprises a dataset of 8,500 diverse and high-quality math word problems at the grade school level, designed to challenge LLMs with tasks necessitating advanced, multi-step reasoning abilities. GSM8K's primary aim is to gauge how well these models can parse, understand, and solve math problems, thereby offering a comprehensive measure of their capacity for logical reasoning and mathematical computation. By incorporating such a specialized benchmark, researchers can better understand the extent to which LLMs can mimic human-like reasoning in solving complex mathematical scenarios. For evaluation on GSM8K, we do not use specific prompts and follow the original test prompts.

\begin{figure}[t]
    \centering
    \begin{minipage}{0.45\textwidth}
        \centering
        \includegraphics[width=0.9\textwidth]{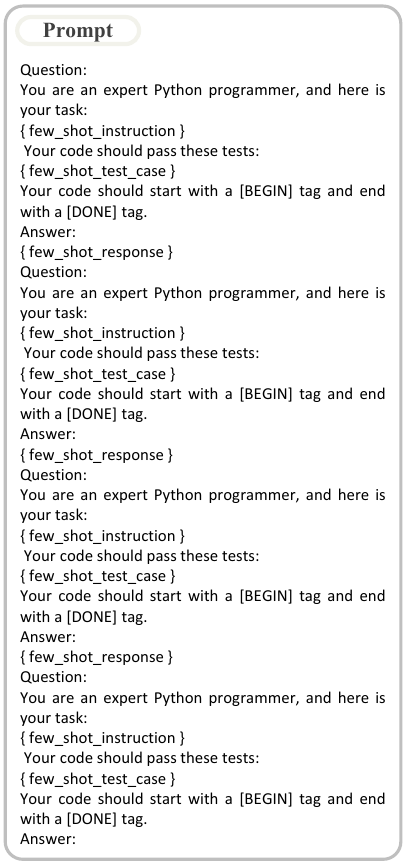} % Reduce the figure size so that it is slightly narrower than the column.
        \caption{Three-shot prompt used to evaluate on MBPP and MBPP+.}
        \label{figure: MBPP Evaluation Prompt}
    \end{minipage}
    \hfill
    \begin{minipage}{0.53\textwidth}
        \centering
        \includegraphics[width=0.97\textwidth]{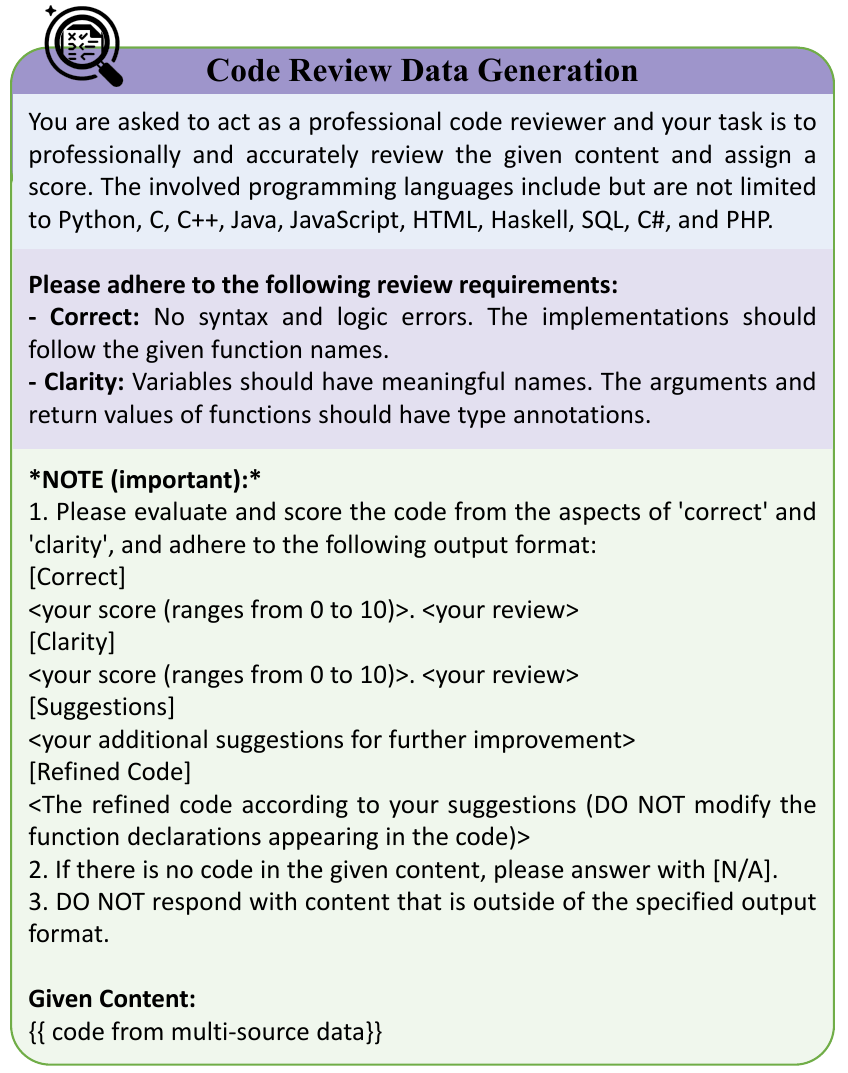} % Reduce the figure size so that it is slightly narrower than the column.
        \caption{Detailed prompt designed for generating code review data.}
        \label{figure: Code Review Data Generation}
    \end{minipage}
\vspace{-12pt}
\end{figure}

\section{AlchemistCoder Dataset Details}\label{AlchemistCoder Dataset Details}
\subsection{\promptname{}}
\begin{figure*}[]
\centering
\includegraphics[width=0.75\textwidth]{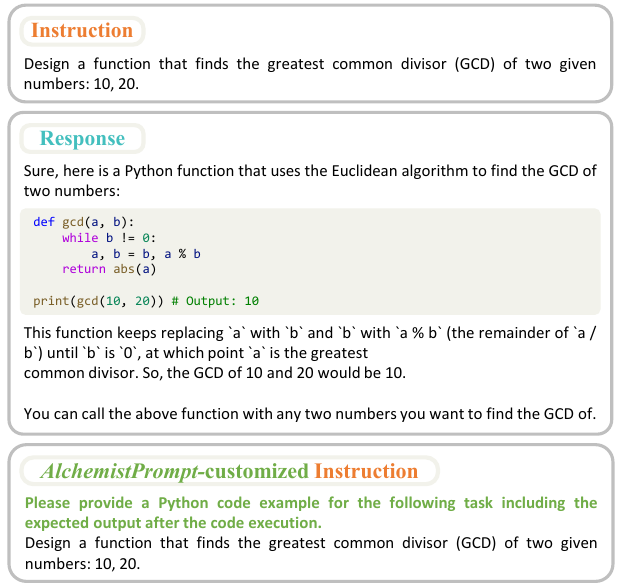} % Reduce the figure size so that it is slightly narrower than the column.
\caption{Example \#1 of \promptname{}s.}
\label{figure: ALCHEMIST-PROMPT Case 1}
\end{figure*}

\begin{figure*}[]
\centering
\includegraphics[width=0.75\textwidth]{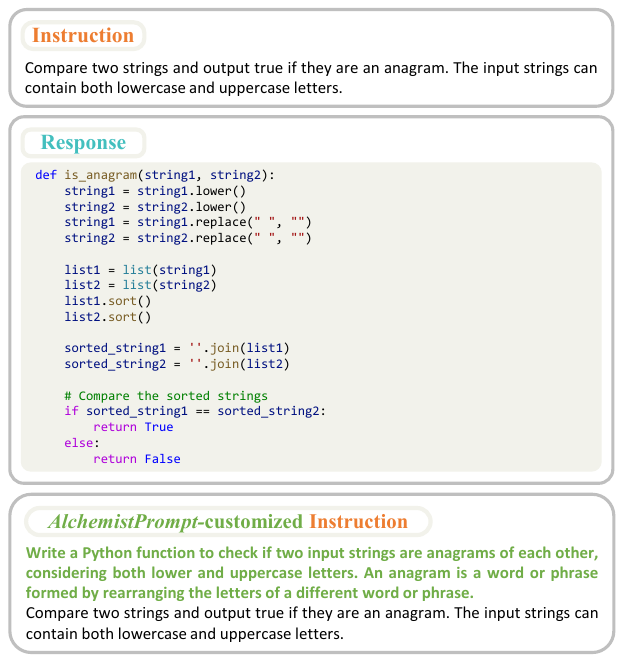} % Reduce the figure size so that it is slightly narrower than the column.
\caption{Example \#2 of \promptname{}s.}
\label{figure: ALCHEMIST-PROMPT Case 2}
\end{figure*}

We provide two samples of \promptname{}s in Fig. \ref{figure: ALCHEMIST-PROMPT Case 1} and Fig. \ref{figure: ALCHEMIST-PROMPT Case 2}.

\subsection{Code comprehension task data}
\subsubsection{Instruction Evolution Data}
\begin{figure*}[]
\centering
\includegraphics[width=0.87\textwidth]{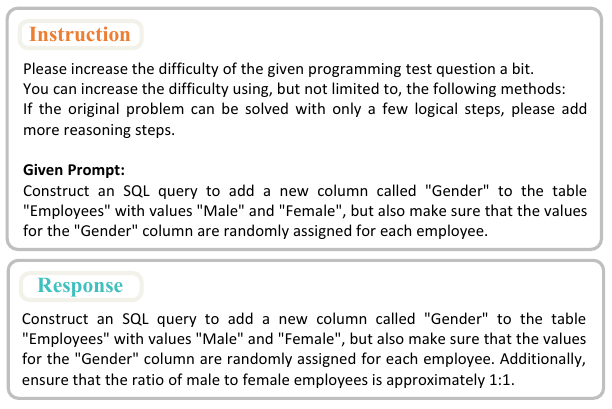} % Reduce the figure size so that it is slightly narrower than the column.
\caption{Example \#1 of instruction evolution task data.}
\label{figure: Instruction Evolution Data Case 1}
\end{figure*}

\begin{figure*}[]
\centering
\includegraphics[width=0.87\textwidth]{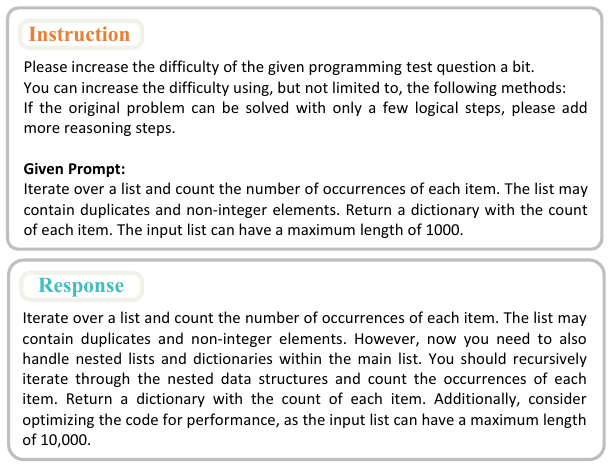} % Reduce the figure size so that it is slightly narrower than the column.
\caption{Example \#2 of instruction evolution task data.}
\label{figure: Instruction Evolution Data Case 2}
\end{figure*}

We provide two samples of instruction evolution task data in Fig. \ref{figure: Instruction Evolution Data Case 1} and Fig. \ref{figure: Instruction Evolution Data Case 2}.

\subsubsection{Data filtering data}
\begin{figure*}[htbp]
\centering
\includegraphics[width=0.87\textwidth]{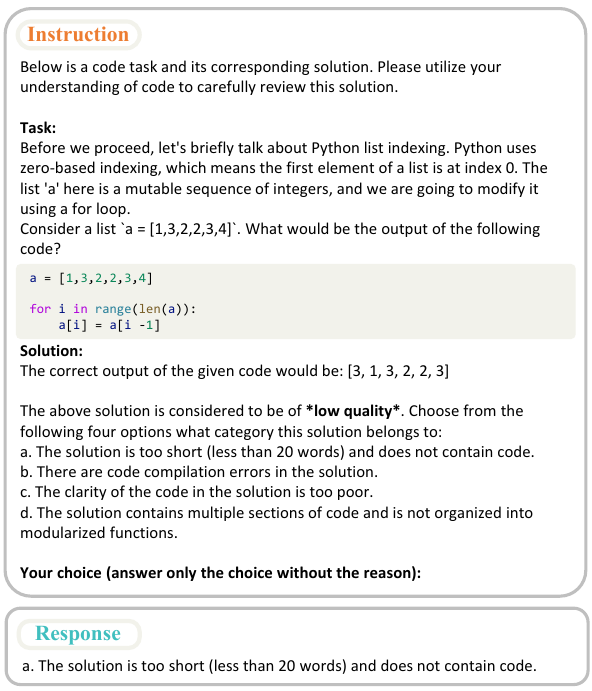} % Reduce the figure size so that it is slightly narrower than the column.
\caption{Example \#1 of data filtering task data.}
\label{figure: Data Filtering Data Case 1}
\end{figure*}

\begin{figure*}[htbp]
\centering
\includegraphics[width=0.87\textwidth]{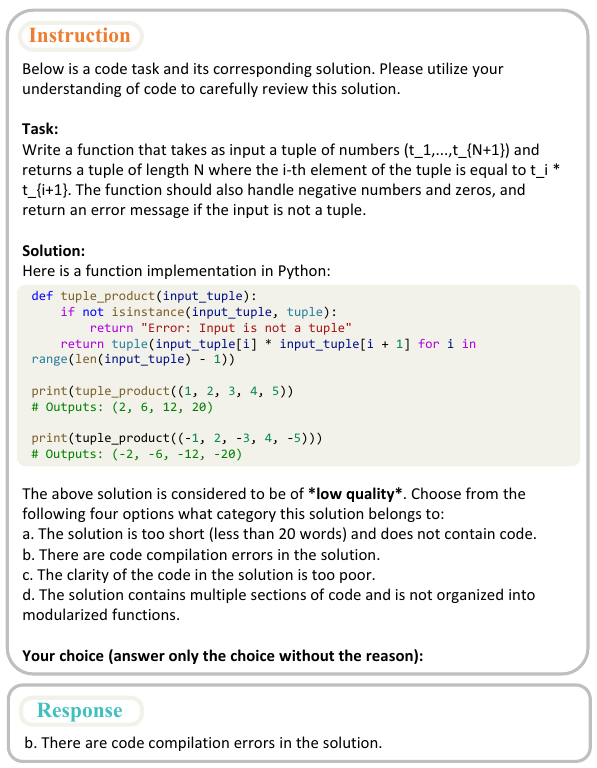} % Reduce the figure size so that it is slightly narrower than the column.
\caption{Example \#2 of data filtering task data.}
\label{figure: Data Filtering Data Case 2}
\end{figure*}

We provide two samples of data filtering task data in Fig. \ref{figure: Data Filtering Data Case 1} and Fig. \ref{figure: Data Filtering Data Case 2}.

\subsubsection{Code review data}
\begin{figure*}[htbp]
\centering
\includegraphics[width=0.87\textwidth]{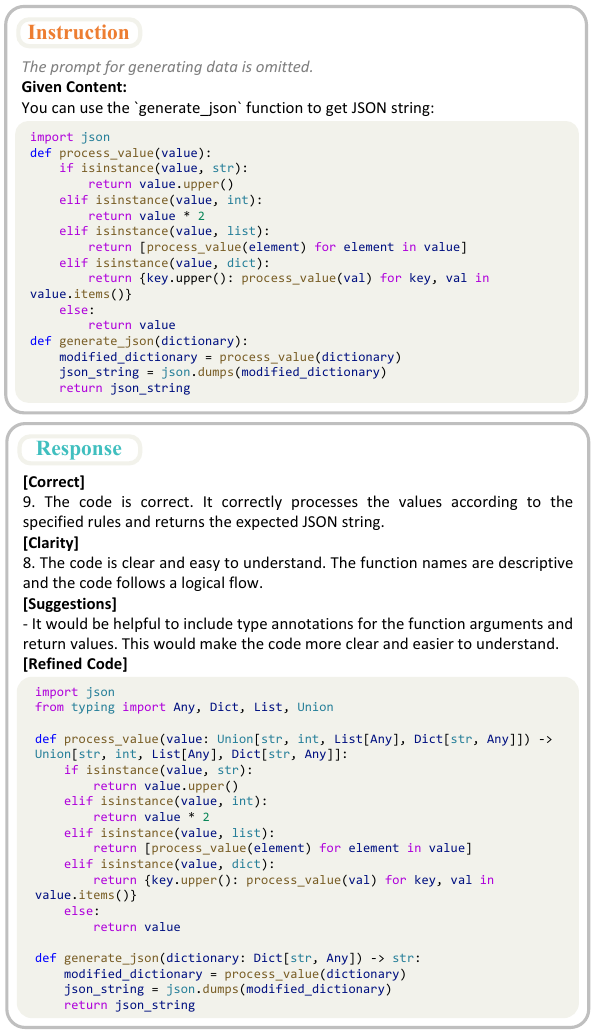} % Reduce the figure size so that it is slightly narrower than the column.
\caption{Example \#1 of code review task data. For clarity, the prompt for generating data is omitted.}
\label{figure: Code Review Data Good Case}
\end{figure*}

\begin{figure*}[htbp]
\centering
\includegraphics[width=0.87\textwidth]{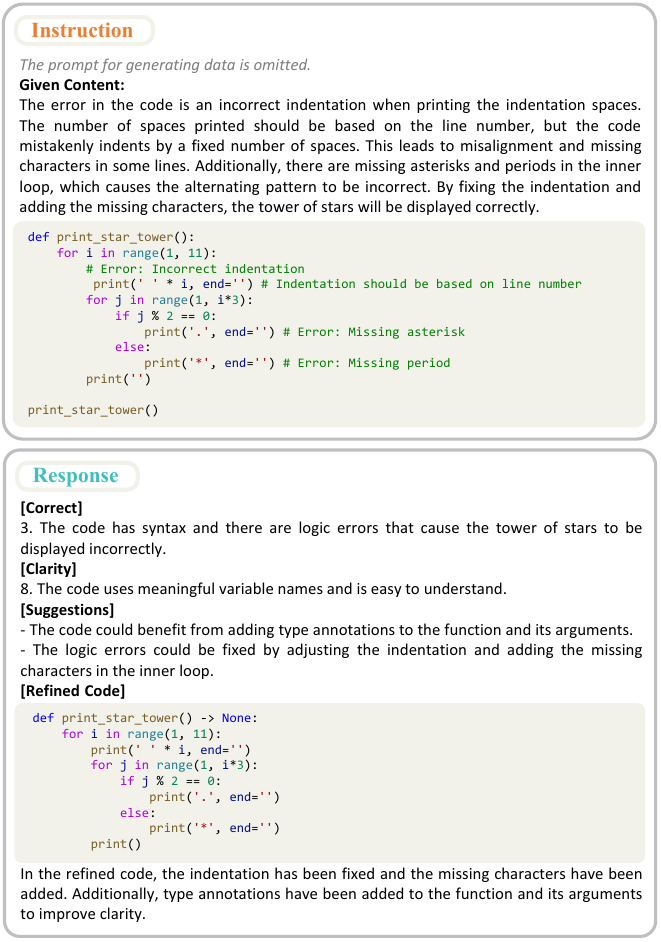} % Reduce the figure size so that it is slightly narrower than the column.
\caption{Example \#2 of code review task data. For clarity, the prompt for generating data is omitted.}
\label{figure: Code Review Data Bad Case}
\end{figure*}

We design prompt as illustrated in Fig. \ref{figure: Code Review Data Generation} to obtain high-quality code review task data and we provide two samples in Fig. \ref{figure: Code Review Data Good Case} and Fig. \ref{figure: Code Review Data Bad Case}.

\newpage

\newpage

\end{document}